# A Global Transport Capacity Risk Prediction Method for Rail Transit Based on Gaussian Bayesian Network

Zhengyang Zhang[1]
[1]Department of Automation, Tsinghua University,
Beijing, China
e-mail: zhengyan20@mails.tsinghua.edu.cn

*Wei Dong[2]
[2]Beijing National Research Center for Information Science and Technology, Tsinghua University, Beijing, China
*e-mail: weidong@mail.tsinghua.edu.cn

Jun Liu[3]
[3]Beijing National Railway Research & Design Institute of Signal & Communication Group Co., Ltd
junliu@crscd.com.cn

Xinya Sun[1]
[1]Department of Automation, Tsinghua University,
Beijing, China
e-mail: xinyasun@mail.tsinghua.edu.cn

Yindong Ji[1]
[1]Department of Automation, Tsinghua University,
Beijing, China
e-mail: jyd@mail.tsinghua.edu.cn

*Abstract*—Aiming at the prediction problem of transport capacity risk caused by the mismatch between the carrying capacity of rail transit network and passenger flow demand, this paper proposes an explainable prediction method of rail transit network transport capacity risk based on linear Gaussian Bayesian network. This method obtains the training data of the prediction model based on the simulation model of the rail transit system with a three-layer structure including rail transit network, train flow and passenger flow. A Bayesian network structure construction method based on the topology of the rail transit network is proposed, and the MLE (Maximum Likelihood Estimation) method is used to realize the parameter learning of the Bayesian network. Finally, the effectiveness of the proposed method is verified by simulation examples.

## I. INTRODUCTION

Rail transit plays an increasingly important role in modern urban transportation with its advantages of large capacity, good punctuality, high safety, environmental friendliness and low cost, and has become the backbone and important support of modern transportation. Although the safety of rail transit is higher than that of conventional road traffic, due to the large scale of rail transit network, heavy transportation tasks and close coupling between lines, once a failure or safety accident occurs, it will have a great impact on urban transportation. For example, on December 22, 2009, around 7:00 a.m., a collision occurred on Shanghai Metro Line 1, which caused a 4-hour traffic paralysis due to inadequate disposal measures and brought serious impact to the travel of a large number of passengers[1]. Therefore, how to effectively evaluate and predict the overall risk of rail transit network and reduce the overall risk and its impact on the transport capacity of rail transit network has always been one of the key issues in the field of rail transit.

In our preliminary work [1-2], we proposed the concept of transport capacity risk of rail transit network. Transport capacity risk takes the carrying capacity of rail transit network for transportation demand as the core, and considers the inherent relationship and mutual influence of transport capacity risk and traditional single-point risk (i.e., equipment risk, personnel risk, environmental risk, and so on). On the one hand, it can reflect the matching of supply and demand of rail transit network transport capacity, and on the other hand, it can represent the influence of various traditional single-point risks on the global transport capacity of the network. As a result, transport capacity risk can realize the effective assessment of the global risk of the rail transportation network with transport capacity as the core under the conditions of reasonableness, completeness and compatibility. However, traditional single-point risk indicators often focus on the economic loss of personnel or equipment caused by events such as equipment damage and vehicle outage, and cannot assess the impact of risk sources on the matching of supply and demand of the transportation network.

Since transport capacity risks at the rail transit network level have a large influence surface and propagation inertia, different passenger flow conditions will also have different impacts on the safety of the network, if effective preventive measures are not taken, once the risk propagation starts, it can easily lead to a rapid decline in the safety of the whole network and eventually lead to safety accidents. Therefore, the prediction of transport capacity risk on the basis of transport capacity risk assessment has important practical significance for the safe operation of rail transit network.

## II. RELATED WORKS

Risk assessment has been one of the important research contents in the field of rail transportation because passenger transport is concerned with safety of passenger groups and important equipment assets. In this regard, Tiankun Xu[3] established the Interpretative Structural Model (ISM) of operational accident influencing factors according to the characteristics of urban rail transit operational accidents and risk influencing factors, and identified the causal mechanisms of operational safety risk factors and the complex correlations among the factors. Chao Wu [4] applied complex network theory to urban rail transit safety and accident analysis, and introduced Fuzzy Relation Analysis (FRA) method to estimate the risk level of different risk factors or events. Among them, the fuzzy hierarchical analysis method was used to compare the relative importance of each risk factor or event. Wang Xu [5] combined fuzzy set theory and Bayesian network method to evaluate the safety of urban rail transit operation from four influencing factors of man-equipment-environment-management. Szymula et al. [6] studied the performance and behavior of rail transit network under the risk of disruption and construct a rail transit network vulnerability assessment model from the perspective of passenger flow and train scheduling, revealing the relationship among critical path, passenger flow demand and static topology of the network under the risk situation. Jiateng Yin [7] et al. constructed a quantitative resilience assessment model of urban rail transit system based on Bayesian network for Beijing metro fault data, and the potential causal relationships among the risk factors of the rail transit system and the weaknesses of the system can be explored by this model. Existing research on rail transit risk assessment mainly focus on equipment and facility failures and traditional risk factors such as man-equipment-environment-management, but less consideration is given to the global transport capacity risk at the transportation level and the interaction between the global transport capacity risk and local single-point risk.

Transport capacity is the core capacity of rail transportation system, and therefore the assessment and enhancement of rail transit transport capacity is also an important research content in the field of rail transportation. In this respect, Zhongli Lei [8] provided a definition of the transport capacity of the rail transit network and proposed a probability-based methodology for calculating the transport capacity reliability of the rail transit network by incorporating the stochastic components in the actual operation of the rail transit network. Yajing Zheng [9] further considered the impact on rail transit network transport capacity when external conditions such as rail transit network structure and transport demand changed, proposed the concept of rail transit network capacity adaptability, analyzed the cascading failure effect after rail transit network damage, and proposed a series of indicators of rail transit network transport capacity resistance to destruction. The research in this area mainly focus on the assessment of the transport capacity itself or the reliability assessment of the network structure and topology, but there are few researches considering the global transport capacity and the supply and demand matching of the network, as well as the interaction between the single point risk and the transport capacity risk.

In terms of rail transit transport capacity risk assessment and prediction, related research has just started. Mengyu Zhang [1] firstly constructed a global RAMS assessment system with transport capacity risk as the core in the regional rail transit scenario from the relationship between network transport capacity and passenger demand load. However, this work can only assess static transport capacity risk and does not involve transport capacity risk prediction. Wenchao Cui [2] extended the static transport capacity risk assessment system proposed by Zhang [1] and defined dynamic transport capacity risk assessment indexes based on the analysis of regional rail transit passenger flow lines and dynamic passenger flow variation. The global dynamic transport capacity risk was evaluated, combined with factors such as the dissipation time of passengers in the waiting area. And an SVM-based dynamic transport capacity risk prediction method was proposed based on transport capacity risk feature extraction. The main shortcoming of this work is that the transport capacity risk prediction is entirely data-driven, therefore the model lacks interpretability, and the model cannot be revised and extended based on expert knowledge.

In summary, the main shortcomings of the existing research on rail transit risk assessment and transport capacity assessment are: 1) The interaction between single point risk and transport capacity of the network is less considered, thus the global risk and global transport capacity of rail transportation are not assessed as a whole; 2) In regard to transport capacity reliability, more consideration is given to the reliability of the network structure and topology, and there is a lack of research on the global transport capacity risk based on matching the global dynamic transport capacity of the network with the dynamic passenger flow demand. In terms of rail transit transport capacity risk prediction, the models established by existing methods lack interpretability and cannot be modified and extended based on expert knowledge. Therefore, based on the previous work, this paper proposes a Bayesian network-based dynamic transport capacity risk prediction method for rail transit, and establishes an explainable transport capacity risk prediction model to provide a basis for further global risk situational awareness and integrated decision support for rail transit.

III. RAIL TRANSIT TRANSPORT CAPACITY RISK ASSESSMENT SYSTEM

This paper adopts the rail transit transport capacity risk assessment system proposed by Zhang [1] and Cui [2], and the relevant contents are briefly described as follows.

*A. Design principles and overall ideas*

The rail transit transport capacity risk indexes are used to assess the global carrying capacity of the rail transit network to the passenger flow OD demand, which must primarily satisfy the following design principles [1]:

1) Reasonability: Reasonability refers to the logical and explainable relationship between index results and their influencing factors, which can correctly reflect the influence of the change of influencing factors on the final index.

2) Completeness: Completeness means that when establishing the index system, the influencing factors of the indexes should be analyzed and considered comprehensively, and the relevant influencing factors should not be omitted.

3) Compatibility: Compatibility means that the newly established rail transit transport capacity risk indicator system should not conflict with the existing related rail transit risk and capacity indicator system.

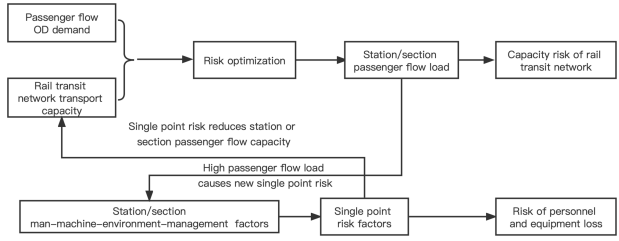

Figure 1.  Static transport capacity risk assessment system for rail transit

Based on the above principles, the literature [1] gives the overall idea and structural design of the static transport capacity risk index system for rail transportation, as shown in Fig. 1. Figure 1 is briefly described as follows:

1) The core element of rail transit transport capacity risk is the matching relationship between OD passenger demand and network capacity. This matching relationship is affected by risk optimization measures such as train scheduling and passenger guidance, which in turn forms the passenger load (i.e., the ratio of passenger demand to passenger capacity[1]) of each station and section of the network, from which the global passenger demand load risk of the network can be calculated (i.e., the global transport capacity risk of the network).

2) This assessment system not only considers the transport capacity risk, but also naturally integrates the traditional single point risk into it. Specifically, the risk factors such as "man-equipment-environment-management" will cause the traditional "single point risk", which on one hand will affect the function of rail transit equipment and facilities, thus reducing the passenger flow capacity of stations or sections and thus affecting the transport capacity risk, and on the other hand will also cause the loss of personnel and equipment, forming the loss risk of personnel and equipment.

3) Meanwhile, the high load passenger flow at stations or sections also tends to trigger new single-point risks, thus form the interaction between single-point risks and transport capacity risks.

On the basis of the above static transport capacity risk assessment system, the literature [2] further proposed a dynamic transport capacity risk assessment system. The static transport capacity risk assessment system assumes that the passenger flow OD demand has been given, but this condition is difficult to be satisfied in actual application. The dynamic transport capacity risk assessment can only be made based on the passenger flow monitoring information. The main difference between the static and dynamic transport capacity risk assessment is that the actual passenger load obtained from passenger flow monitoring will not exceed 100%, while more incompressible passenger demand will be embodied in the form of stranded passenger flow at stations.

*B. Key Features*

Based on the above overall idea, this section mainly gives the key characteristic used to calculate the global dynamic transport capacity risk of the rail transit network, as shown in Tab. I.

*C. evaluation methodology*

In this paper, the evaluation and calculation method given in [2] is used to calculate the line and global transport capacity risk

TABLE I.  KEY FEATURES OF GLOBAL DYNAMIC TRANSPORT CAPACITY RISK CALCULATION [2]

| Concept | Meaning |
|---|---|
| passenger flow OD demand | $Q$, $q_{ij} \in Q$ represent the passenger flow demand from station $i$ to station $j$, unit: person per hour. |
| actual passenger flow | Actual passenger flow through the station or section within a certain period of time, unit: person per hour. |
| station/ section transport capacity | The maximum number of passengers who can safely pass through the station or section in unit time, unit: person per hour. |
| station/section passenger flow saturation | The ratio of the actual passenger flow through the station or section to the passenger flow capacity of the station or section. It reflects the relative relationship between the actual passenger flow and the passenger flow capacity, and is one of the determinants of the risk probability of the station or section capacity. |
| number of passengers in the waiting area | The number of passengers stranded in the waiting area for a certain period of time. |
| capacity of dissipation path for stranded passenger | The estimated path capacity for the dissipation of stranded passenger at the station. |
| dissipation time of stranded passenger in station | The time required for the stranded passengers in station to leave the station completely. |

corresponding to the rail transit network simulation model, so as to generate the training data and test data of the prediction model. The evaluation calculation method in [2] is briefly described as follows:

According to the principle of additivity of risk loss, the global dynamic transport capacity risk of rail transit network at time t is the sum of the dynamic transport capacity risk of each risk point at that time. According to the risk assessment principle [12], the risk probability of the risk point is multiplied by the risk consequence to obtain the risk value of the risk point. The basic risk points include station passenger flow saturation risk, section passenger flow saturation risk and station stranded passenger detention risk.

On this basis, the station transport capacity risk, line transport capacity risk and global transport capacity risk of rail transit network can be further calculated. The station transport capacity risk $RS_i(t)$ is obtained by adding station saturation risk and station passenger detention risk, which indicates the risk value of the $i$th station at time $t$. $RI_j(t)$ is the transport capacity risk of the $j$th section at time $t$. $SS_i(t)$, $SW_i(t)$ and $SI_j(t)$ represent respectively the saturation, the stranded passenger dissipation time of the $i$th station and the saturation of the $j$th section at time $t$, and they are obtained by dividing the station passenger flow $PS_i(t)$, the number of stranded passengers $PW_i(t)$, and the section passenger flow $PI_j(t)$ correspondingly by the station passenger capacity $CS_i(t)$, the station passenger dissipation speed $CW_i(t)$, and the section passenger capacity $CI_j(t)$, respectively. In addition, $w1_i(t), w2_i(t), w3_j(t)$ are the risk consequence of station passenger flow saturation, station passenger detention, and section passenger flow saturation respectively.

---

[1] This is not the ratio of actual passenger flow to passenger capacity, so the passenger load can be greater than 100%, which is the main difference from the "dynamic transport capacity risk assessment system" mentioned later.

The line transport capacity risk $RL_k(t)$ is obtained by adding up the transport capacity risk of all the stations and sections within the line, representing the risk value of the $k$th line, while the network transport capacity risk $RN(t)$ is derived by summing up the risk of all stations and sections in the network $N$, as in (1)-(7).

$$RS_i(t) = w1_i(t) * f(SS_i(t)) + w2_i(t) * f(SW_i(t)) \quad (1)$$

$$RI_j(t) = w3_j(t) * f(SI_j(t)) \quad (2)$$

$$RL_k(t) = \sum_{i \in k} RS_i(t) + \sum_{j \in k} RI_j(t) \quad (3)$$

$$RN(t) = \sum_{i \in N} RS_i(t) + \sum_{j \in N} RI_j(t) \quad (4)$$

$$SS_i(t) = \frac{PS_i(t)}{CS_i(t)} \quad (5)$$

$$SW_i(t) = \frac{PW_i(t)}{CW_i(t)} \quad (6)$$

$$SI_j(t) = \frac{PI_j(t)}{CI_j(t)} \quad (7)$$

The expression for $f(x)$ given in [1] is adopted, which is shown in (8). The corresponding parameters a and b are 6 and 7, respectively.

$$f(x) = \frac{1}{1 + e^{-ax+b}} \quad (8)$$

If a risk event occurs in an urban rail system, the severity of the consequences increases with the number of people gathered. Therefore, the value of $w1_i(t), w2_i(t), w3_j(t)$ should be chosen according to the magnitude of the passenger flow at the stations and sections, see the literature [2] for details.

## IV. GLOBAL DYNAMIC RISK PREDICTION METHOD FOR RAIL TRANSIT NETWORK BASED ON LGBN

### A. Linear Gaussian Bayesian network

Bayesian network is a probabilistic graphical model proposed by Judea Pearl [11], which can simulate the uncertainty of causality in the reasoning process. It is a directed acyclic graph (DAG) composed of nodes representing variables and directed edges connecting these nodes. The relationship between nodes in discrete BN can be linear or nonlinear and can be used for complex model construction, so it is widely used in rail transit risk assessment [12-14]. However, the discretization of continuous variables will lead to the loss of a large amount of feature information, and continuous Bayesian networks can avoid this problem. GBNs[2] are a simple and efficient class of continuous Bayesian networks, and the joint probability distribution of a GBN is a multivariate normal distribution whose Conditional Probability Distribution, CPD) is linearly Gaussian [15]. In the rail transit network transport capacity risk assessment system, there is a strong linear relationship between the transport capacity risks of different levels, such as the line risk and the station risk, so the linear hypothesis is applicable to the current scenario.

### B. Bayesian network construction based on rail transit network topology

In practice, rail transit network transport capacity risk is closely related to the structure of rail transit network. Therefore, we propose a Bayesian network construction method based on rail transit network topology. From the perspective of topology, the line is composed of stations and sections between stations. The carrying capacity of the line is more directly reflected by the passenger flow throughput level of stations and sections and the passenger retention level of stations. The former is measured by passenger flow saturation, and the latter is measured by the evacuation time of the stranded passengers.

Since the line risk is the summation of the risk of each station and section within the line, and the global risk of the rail transit network is obtained by adding up the risk of all stations and sections in the network, the overall structure of the Bayesian network is designed as Fig.2. In the Bayesian network constructed in this paper, the directed connection relationship between nodes represents the causal relationship between nodes. Specifically, the nodes representing the station passenger flow saturation and the station stranded passenger dissipation time directly point to the corresponding station risk nodes, while the nodes representing the section passenger flow saturation point to the section risk nodes. All the "station risk" and "section risk" nodes point to the "line risk" node representing the line where the stations/sections are located. As a result, those nodes belonging to transfer stations or repeat sections point to multiple "line risk" nodes. Finally, all the "line risk" nodes point to the "global risk" node. It should be noted that, according to the calculation method given in Section III-C, the network global transport capacity risk is not equal to the sum of all lines' transport capacity risk due to the existence of interchange nodes and repetition sections. The weight value of different nodes to the output transport capacity risk can be used to determine which nodes have a greater impact on the results and thus to perform the reverse analysis. By constructing a risk prediction network from topological relationships, it is possible to predict the global risk of the network, and also to sense the local risk of the network dynamically for the weak points of the network.

---

[2] GBNs are also called linear Gaussian Bayesian networks (LGBNs) because the conditional dependencies of a Gaussian Bayesian network (GBN) are linear.

Tab. II and III gives the basic construction of the Bayesian networks for transport capacity risk prediction of Chongqing rail transit network, and the global transport capacity risk prediction hierarchy of rail transit network is shown in Fig. 2. The variables represented by each node of the Bayesian network and their hierarchical descriptions are shown in Tab. IV.

### C. Bayesian network parameter learning for transport capacity risk prediction

The purpose of Bayesian network parameter learning is to obtain conditional dependencies between risk nodes at different levels. In the case of a given data set, the maximum likelihood estimation (MLE) method is usually used to estimate the parameters of GBN.

For a GBN, whose joint probability density is a multivariate normal distribution[17], for any node Y in the network, let $\mathbf{X}$ consist of all the random variables of its parents, i.e. $\mathbf{X} = (X_1, X_2, \ldots, X_k)$, $X_k \in Pa(Y)$, then the conditional probability $P(Y|\mathbf{X})$ obeys a linear Gaussian distribution $N(\mu_{Y|X}; \sigma^2)$, whose parameters satisfy that

$$\mu_{Y|X} = \beta_0 + \beta_1 X_1 + \beta_2 X_2 + \cdots + \beta_k X_k \qquad (11)$$

In Bayesian network parameter learning, we want to obtain the corresponding parameters of the conditional probability distribution, and the goal is to find the parameter estimate $\theta_{Y|X} = <\beta_0, \ldots, \beta_k, \sigma^2>$ for a given data set, i.e. ($\theta_{Y|X}: \mathcal{D}$), so we estimate the parameters in the Bayesian network using the great likelihood estimation method on the training set $\mathcal{D}$. In order to find the value of the great likelihood parameters, the likelihood function needs to be differentiated and the equations needs to be solved. To facilitate the calculation, the log-likelihood function is used, and according to the definition of Gaussian distribution, the corresponding log-likelihood function can be written as follows[17]:

$$\ell_Y(\theta_{Y|X}: \mathcal{D}) = \log L_Y(\theta_{Y|X}: \mathcal{D})$$
$$= \sum_m \left[-\frac{1}{2}\log(2\pi\sigma^2) - \frac{1}{2}\frac{1}{\sigma^2}(\beta_0 + \beta_1 x_1[m] + \beta_2 x_2[m] + \cdots + \beta_k x_k[m] - y[m])^2\right] \qquad (12)$$

In equation (12) $m = 1, 2, \ldots, M$, M is the total number of samples in the data set $\mathcal{D}$. Finding the gradient of the above equation to $\beta_0$ yields equation (13).

$$\frac{\partial}{\partial \beta_0}\ell_Y(\theta_{Y|X}: \mathcal{D}) = \sum_m -\frac{1}{\sigma^2}(\beta_0 + \beta_1 x_1[m] + \beta_2 x_2[m] + \cdots + \beta_k x_k[m] - y[m])$$
$$= -\frac{1}{\sigma^2}(m\beta_0 + \beta_1 \sum_m x_1[m] + \beta_2 \sum_m x_2[m] + \cdots + \beta_k \sum_m x_k[m] - \sum_m y[m]) \qquad (13)$$

Let the gradient of the above equation take the value of 0, and we get equation (14).

$$\frac{1}{M}\sum_m y[m] = \beta_0 + \beta_1 \frac{1}{M}\sum_m x_1[m] + \beta_2 \frac{1}{M}\sum_m x_2[m] + \cdots + \beta_k \frac{1}{M}\sum_m x_k[m] \qquad (14)$$

Finding the gradient of equation (12) to $\beta_i$ with $\frac{\partial}{\partial \beta_i}\ell_Y(\theta_{Y|X}: \mathcal{D}) = 0$ yields equation (15).

$$\frac{1}{M}\sum_m y[m] x_i[m] = \beta_0 \frac{1}{M}\sum_m x_i[m] + \beta_1 \frac{1}{M}\sum_m x_1[m] \cdot x_i[m] + \beta_2 \frac{1}{M}\sum_m x_2[m] \cdot x_i[m] + \cdots + \beta_k \frac{1}{M}\sum_m x_k[m] \cdot x_i[m] \qquad (15)$$

Equation (15) represents a total of k equations, which can be solved for k+1 variables $\beta_0, \beta_1, \ldots, \beta_k$ through a system of k+1 linear equations by associating with equation (13), denoted as $\boldsymbol{\beta} = (\beta_1, \ldots, \beta_k)^T$, and the solution result is shown in (16)(17), where $\mu_{Y_\mathcal{D}}, \mu_{X_\mathcal{D}}$ represent the sample means of the variables Y, $\mathbf{X}$ according to data set $\mathcal{D}$. $\Sigma_{Y_\mathcal{D} X_\mathcal{D}}$ is the sample covariance of Y, X in $\mathcal{D}$, $\Sigma_{X_\mathcal{D} X_\mathcal{D}}$ is the sample variance of X in $\mathcal{D}$.

$$\boldsymbol{\beta} = \Sigma_{X_\mathcal{D} X_\mathcal{D}}^{-1} \Sigma_{Y_\mathcal{D} X_\mathcal{D}} \qquad (16)$$

$$\beta_0 = \mu_{Y_\mathcal{D}} - \Sigma_{Y_\mathcal{D} X_\mathcal{D}} \Sigma_{X_\mathcal{D} X_\mathcal{D}}^{-1} \mu_{X_\mathcal{D}} \qquad (17)$$

Similarly, equation (12) is solved for the gradient of $\sigma^2$ and the equation is constructed to obtain equation (18), where $\Sigma_{Y_\mathcal{D} Y_\mathcal{D}}$ is the sample variance of Y in $\mathcal{D}$.

$$\sigma^2 = \Sigma_{Y_\mathcal{D} Y_\mathcal{D}} - \Sigma_{Y_\mathcal{D} X_\mathcal{D}} \Sigma_{X_\mathcal{D} X_\mathcal{D}}^{-1} \Sigma_{X_\mathcal{D} Y_\mathcal{D}} \qquad (18)$$

Although (16)-(18) provide exact solutions for solving linear equations of maximum likelihood estimation, the solutions of the equations may be negative in practical applications, and these negative solutions lack interpretability. In this case, it is more meaningful to use the non-negative least squares (NNLS) method to solve the equations[19]. Therefore, we transform the problem of solving linear equations into an optimization problem, as shown below, and use the **nnls** method of the **sklearn** package in Python to solve this problem.

$$\begin{array}{l} \arg\min_{\boldsymbol{\beta}} \quad \|A\boldsymbol{\beta} - y\|_2^2 \\ s.t \quad \boldsymbol{\beta} \geq \mathbf{0} \end{array} \qquad (19)$$

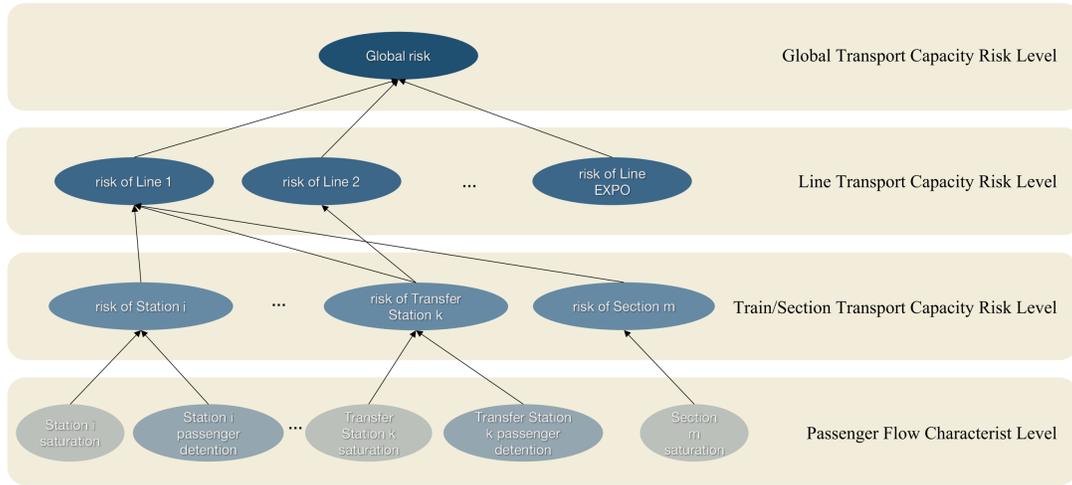

Figure 2. Rail transit network global transport capacity risk prediction Bayesian network hierarchy.

TABLE II. BAYESIAN NETWORK STRUCTURE FOR LINE TRANSPORT CAPACITY RISK PREDICTION

| BN Number | Corresponding Line | The number of non-transfer station nodes | The number of section nodes | The number of transfer station nodes |
|---|---|---|---|---|
| 1 | Chongqing Rail Transit Line 1 | 19 | 46 | 5 |
| 2 | Chongqing Rail Transit Line 2 | 20 | 48 | 5 |
| 3 | Chongqing Rail Transit Line 3 | 31 | 78 | 8 |
| 4 | Chongqing Rail Transit Line 3 (Kong Gang) | 6 | 12 | 1 |
| 5 | Chongqing Rail Transit Line 4 | 6 | 14 | 2 |
| 6 | Chongqing Rail Transit Line 5 | 8 | 18 | 2 |
| 7 | Chongqing Rail Transit Line 6 | 20 | 54 | 8 |
| 8 | Chongqing Rail Transit Line 10 | 14 | 36 | 5 |
| 9 | Chongqing Rail Transit Loop Line | 18 | 50 | 8 |
| 10 | Chongqing Rail Transit International Expo Line | 4 | 10 | 2 |

TABLE III. BAYESIAN NETWORK STRUCTURE FOR GLOBAL TRANSPORT CAPACITY RISK PREDICTION

| BN Number | Prediction Target | The number of line nodes |
|---|---|---|
| 11 | Global transport capacity risk | 10 |

TABLE IV. VARIABLE DESCRIPTION OF GLOBAL TRANSPORT CAPACITY RISK PREDICTION BAYESIAN NETWORK FOR RAIL TRANSIT NETWORK

| Level | Variable Name | Variable Sign | Variable Description | Variable Type |
|---|---|---|---|---|
| Passenger flow characteristic layer | Station passenger flow saturation | $SS_i(t), i = 1, 2, ..., N$ | The ratio of passenger flow through the station to the station's capacity, N is the total number of stations. | Input |
| | Dissipation time of stranded passenger in station | $SW_i(t), i = 1, 2, ..., N$ | The time for passengers stranded in the waiting area of the station to leave the waiting area completely. | Input |
| | Section passenger flow saturation | $SI_j(t), j = 1, 2, ..., M$ | The ratio of passenger flow through the section to the section capacity, M is the total number of sections. | Input |
| Station/section transport capacity risk layer | Station transport capacity risk | $RS_i(t), i = 1, 2, ..., N$ | The transport capacity risk assessment value of stations. | Intermediate |
| | Section transport capacity risk | $RI_j(t), j = 1, 2, ..., M$ | The transport capacity risk assessment value of sections. | Intermediate |
| Line transport capacity risk layer | Line transport capacity risk | $RL_k(t), k = 1, 2, ..., 10$ | The sum of the transport capacity risk of the stations and sections in the line. | Intermediate |
| Global transport capacity risk layer | Global transport capacity risk | $RN(t)$ | The sum of the transport capacity risk of all the stations and sections in the network. | Output |

## D. Application scenarios and simulation models

This paper takes Chongqing rail transit as an example to evaluate and predict the transport capacity risk. The rail transit network, train flow and passenger flow models are constructed for 168 stations and 362 sections of Chongqing Rail Transit Lines 1, 2, 3, 3 (Kong Gang), 4, 5, 6, 10, Loop line and International Expo line. The simulation environment is constructed by Python programming language, object-oriented modeling method and discrete event system simulation method.

The simulation model uses real line length data and topology structure as well as passenger flow OD data after desensitization based on real passenger flow data. A batch of passenger flow data is generated every 15 minutes, and the corresponding passenger flow characteristics, line risk and global risk are calculated. The daily operation time is 6:30-22:30, a total of 16 hours, and 64 batches of passenger flow OD data are generated every day, and the corresponding characteristics and risk assessment values are calculated according to the index calculation method given in Section III. A total of 1920 batches of passenger flow OD data for 30 days are generated.

Referring to the relevant literature[10], stations are divided into large station, medium station and small station. The passenger flow throughput of large station is 21600 people/hour, the passenger flow throughput of medium station is 14400 people / hour, and the passenger flow throughput of small station is 7200 people/hour.

We used the first 1600 data as the training set and the last 320 data as the test set for cross-validation. Tab. V shows the training results of Bayesian network for Chongqing Rail Transit Line 3 (Kong Gang) transport capacity risk prediction. Fig. 3 briefly shows the training results of global transport capacity risk prediction Bayesian network. It should be noted that the relationship between the means in Tab. V is that

$$\mu_{RL_{Line3(K.G.)}} = \beta_0 + \beta_{RS_1} \cdot \mu_{RS_1} + \cdots + \beta_{RS_7} \cdot \mu_{RS_7} + \beta_{RI_{1-2}} \cdot \mu_{RI_{1-2}} + \cdots + \beta_{RI_{7-6}} \cdot \mu_{RI_{7-6}} \quad (20)$$

TABLE V. TRAINING RESULT OF TRANSPORT CAPACITY RISK PREDICTION BAYESIAN NETWORK FOR CHONGQING RAIL TRANSIT LINE 3(KONG GANG)

| Node Variable (Transport Capacity Risk) | Node Identifier | Mean ($\mu$) | Variance ($\sigma^2$) | Line Risk Coefficient ($\beta$) |
|---|---|---|---|---|
| Line_3 (KongGang) | $RL_{Line3(K.G.)}$ | 2.727 | 1.39 | -0.18($\beta_0$) |
| Station 1 | $RS_1$ | 0.014 | 0.002 | 0 |
| Station 2 | $RS_2$ | 0.013 | 0.001 | 11.2019 |
| Station 3 | $RS_3$ | 0.001 | 0.045 | 1.0277 |
| Station 4 | $RS_4$ | 0.004 | 0.008 | 0.5591 |
| Station 5 | $RS_5$ | 0.027 | 0.008 | 1.2093 |
| Station 6 | $RS_6$ | 0.108 | 0.061 | 0.9392 |
| Station 7 | $RS_7$ | 0.011 | 0.001 | 4.4847 |
| Section 1_2 | $RI_{1-2}$ | 0.002 | 0.04 | 0 |
| Section 2_1 | $RI_{2-1}$ | 0.47 | 0.134 | 1.5339 |
| Section 2_3 | $RI_{2-3}$ | 0.003 | 0.005 | 3.0185 |
| Section 3_2 | $RI_{3-2}$ | 0.488 | 0.139 | 0.6775 |
| Section 3_4 | $RI_{3-4}$ | 0.001 | 0.013 | 0.4559 |
| Section 4_3 | $RI_{4-3}$ | 0.476 | 0.149 | 1.7123 |
| Section 4_5 | $RI_{4-5}$ | 0.001 | 0.028 | 1.8201 |
| Section 5_4 | $RI_{5-4}$ | 0.436 | 0.145 | 0.7887 |
| Section 5_6 | $RI_{5-6}$ | 0.001 | 0.007 | 0.5621 |
| Section 6_5 | $RI_{6-5}$ | 0.243 | 0.098 | 1.2862 |
| Section 6_7 | $RI_{6-7}$ | 0.002 | 0.001 | 6.8034 |
| Section 7_6 | $RI_{7-6}$ | 0.014 | 0.001 | 3.0 |

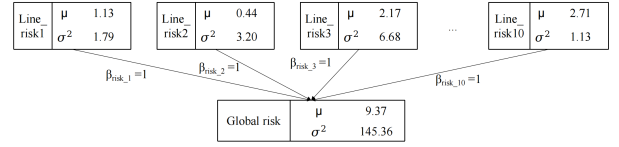

Figure 3. Training result of global transport capacity risk prediction Bayesian network

## V. SIMULATION VERIFICATION RESULTS

According to the simulation verification scenario given in Section. IV-D, we construct a corresponding simulation model to evaluate the transport capacity risk of rail transit network and lines. The Bayesian network structure proposed in IV-B is used here, and the maximum likelihood estimation method is used for parameter learning. Finally, its prediction effect is compared with the autoregressive (**AR**) method as the benchmark scheme.

The benchmarking scheme (**AR**), using only the current and previous values of each transport capacity risk node (station, line or network) itself as input, without considering additional information, predicts the transport capacity risk of the node itself at the next moment (i.e., the input of the node itself at moments $t_0$ and $t_0-\Delta t$ is used to predict the output of the node itself at moments $t_0+\Delta t$, without considering the correlation between the nodes).

In first scheme, a complete Bayesian network is constructed according to the topology of the entire rail transit network (**GBN1**). The inputs at time $t_0$ are used to directly predict the transport capacity risk of each line and the global transport capacity risk of the rail transit network at time t ($t>t_0$)

The second scheme, on the basis of **GBN1**, considers the correlation of prediction data in time series and studies the influence of historical information on prediction effect (**GBN2**). By adding a network input variable at $t0-\Delta t$ time (that is, the input at time $t0$ and $t0-\Delta t$ is used for prediction.), the line transport capacity risk and global transport capacity risk are predicted and compared with the above schemes.

The evaluation index is Weighted Mean Absolute Percentage Error (WMAPE)[16], which considers the weight of each feature on the basis of MAPE[16]. MAPE gives very large error results for smaller prediction errors on features with risk values tending to zero, and this problem can be avoided by using a weighting approach. As shown in (21) and (22), $y_t^*$ is the predicted value of the true risk value $y_t$, and n is the number of samples in the test data set.

$$\text{MAPE} = \frac{1}{n}\sum_{t=1}^{n}\left|\frac{y_t-y_t^*}{y_t}\right| \quad (21)$$

$$\text{WMAPE} = \frac{\sum_{t=1}^{n}(|y_t - y_t^*|)}{\sum_{t=1}^{n}(|y_t|)} \qquad (22)$$

The original global transport capacity risk of Chongqing rail transit network in a certain period of time is evaluated by the simulation model as shown in Fig. 4.

From Fig. 4, it can be seen that the global transport capacity risk has a strong time periodicity, in which the time period represented by the data peak is consistent with the peak period of the daily operation of the rail transit network, representing that the risk events are more likely to occur at this time, which is consistent with the law of the operation state of the rail transit network.

According to the Bayesian network structure proposed in Section IV-B, the passenger flow characteristics serve as the evidential variables to obtain the global transport capacity risk prediction values of each station, each section, each line and rail transit network simultaneously.

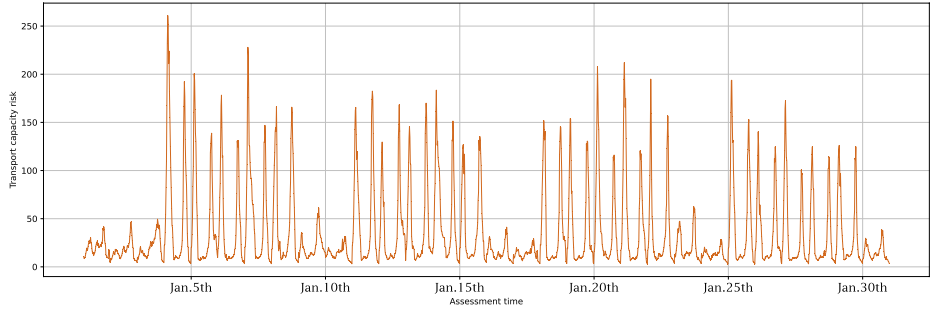

Figure 4. The original global transport capacity risk of Chongqing rail transit network

The result of **GBN1** and **GBN2** are shown in Fig.5 and Fig.6. The above training methods are compared and the results are show in Tab. VI.

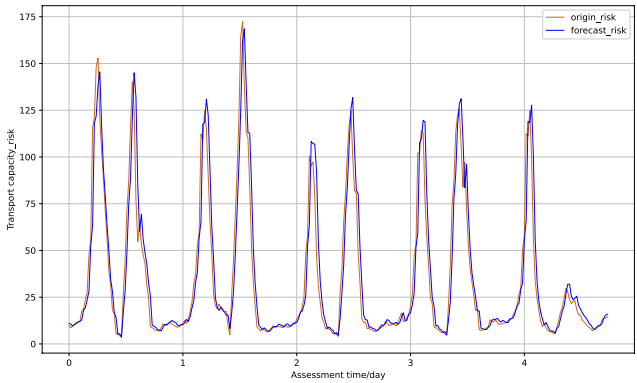

Figure 5. Prediction result of global transport capacity risk in **GBN1**

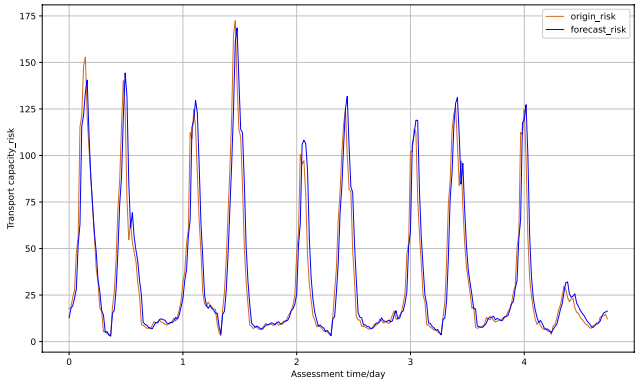

Figure 6. Prediction result of global transport capacity risk in **GBN2**

In terms of WMAPE, the prediction error obtained by using **GBN1** is reduced from 36.88% to 24.44% compared with the AR, indicating the proposed prediction model more satisfactory, which verifies the effectiveness of our proposed method. After adding historical data, the prediction effect is improved significantly (from 24.44% in **GBN1** to 19.10% in **GBN2**), indicating that historical data provides more effective and relevant information for the prediction model. The introduction of historical passenger flow characteristic data can reflect the changing trend of passenger flow and transport capacity risk to a certain extent, improve the prediction of station transport capacity risk and line transport capacity risk, and finally improve the prediction effect of global transport capacity risk.

TABLE VI. TRANSPORT CAPACITY RISK PREDICTION ERROR(WMAPE)

| Target | AR | GBN1 | GBN2 |
|---|---|---|---|
| Line_1 | 49.27% | 37.52% | 37.46% |
| Line_2 | 37.88% | 33.91% | 22.61% |
| Line_3 | 39.88% | 29.34% | 26.43% |
| Line_4 | 28.65% | 24.09% | 24.85% |
| Line_5 | 15.38% | 10.10% | 12.99% |
| Line_6 | 42.13% | 38.61% | 28.08% |
| Line_7 | 37.68% | 26.82% | 20.18% |
| Line_8 | 39.58% | 31.47% | 30.15% |
| Line_9 | 36.50% | 28.37% | 31.08% |
| Line_10 | 33.21% | 30.26% | 24.12% |
| Global | 36.88% | 24.44% | 19.10% |

## VI. CONCLUSION

To address the problem that the existing rail transport capacity risk prediction methods are not explainable, this paper proposes a linear Gaussian Bayesian network based rail transit network transport capacity risk prediction method with interpretability. This paper constructs a Bayesian network for rail transit network transport capacity risk prediction based on the topology of the network, calculates the multi-level transport capacity risk results of the network by establishing a simulation model and combining with the transport capacity risk assessment method to obtain the data set, and uses the great likelihood estimation method to realize the parameter learning of the linear Gaussian Bayesian network. Finally, the validity of the prediction model is verified based on the data related to Chongqing rail transit system, reflecting the role of a priori knowledge such as rail transit network structure in enhancing the prediction effect. The results of the algorithm show that the Bayesian network prediction effect is better than that of simple autoregression, while the prediction effect of Bayesian network with historical data is further improved.

In the next step, we should consider other influencing factors in "man-machine-environment-management" during the actual rail transit operation, expand the prediction network structure, and construct a rail transit safety situational awareness system with transport capacity risk as the core. In addition, the changes of rail traffic passenger flow and transport capacity risk are often time-dependent, therefore, dynamic Bayesian network can be used for research on transport capacity risk prediction and further expand the structure and capability of Bayesian network.

ACKNOWLEDGMENT

This work was supported by the National Key Research and Development Program of China under Grant 2022YFB4300502 and the Research and Development Project of CRSC Research & Design Institute Group Co., Ltd.